\documentclass[review]{elsarticle}
\journal{Information Sciences}
\graphicspath{{img/}}
\DeclareGraphicsExtensions{.pdf,.jpeg,.png}
\usepackage[cmex10]{amsmath}
\usepackage{url}
\usepackage[tight,footnotesize]{subfigure}

\begin{document}
\begin{frontmatter}
\author{A. Ramos-Soto}
\ead{alejandro.ramos@usc.es, alejandro.soto@abdn.ac.uk}
\address{Centro Singular en Investigaci\'on en Tecnolox\'ias da Informaci\'on (CiTIUS), Universidade de Santiago de Compostela, Spain}
\address{Department of Computing Science, University of Aberdeen, United Kingdom}

\author{M. Pereira-Fari\~na}
\ead{martin.pereira@usc.es, m.z.pereirafarina@dundee.ac.uk}
\address{Departamento de Filosof\'ia e Antropolox\'ia, Universidade de Santiago de Compostela, Spain}
\address{Centre for Argument Technology (ARG-tech), University of Dundee, United Kingdom}
\title{On modeling vagueness and uncertainty in data-to-text systems through fuzzy sets}

\begin{abstract}
Vagueness and uncertainty management is counted among one of the challenges that remain unresolved in systems that generate texts from non-linguistic data, known as data-to-text systems. In the last decade, work in fuzzy linguistic summarization and description of data has raised the interest of using fuzzy sets to model and manage the imprecision of human language in data-to-text systems. However, despite some research in this direction, there has not been an actual clear discussion and justification on how fuzzy sets can contribute to data-to-text for modeling vagueness and uncertainty in words and expressions. This paper intends to bridge this gap by answering the following questions: What does vagueness mean in fuzzy sets theory? What does vagueness mean in data-to-text contexts? In what ways can fuzzy sets theory contribute to improve data-to-text systems? What are the challenges that researchers from both disciplines need to address for a successful integration of fuzzy sets into data-to-text systems? In what cases should the use of fuzzy sets be avoided in D2T? For this, we review and discuss the state of the art of vagueness modeling in natural language generation and data-to-text, describe potential and actual usages of fuzzy sets in data-to-text contexts, and provide some additional insights about the engineering of data-to-text systems that make use of fuzzy set-based techniques. 

\end{abstract}


\end{frontmatter}

\section{Introduction} \label{sec:intro}

The vast amounts of data that companies, experts and users need to manage usually appear in very different formats (tables, time-series, images, etc.) and their handling by human users is a real challenge. This has led to the emergence of computational systems that interpret and convert such data into texts, known as natural language generation (NLG) systems. Thus, NLG can be defined as the branch of Artificial Intelligence devoted to research the process of generating information in the form of natural language texts from different types of input data, such as other texts, numeric data or visual information~\cite{nlg_survey}.


Within NLG, systems that use non-linguistic data as input (such as time series data, or numerical datasets in general) are commonly known as data-to-text (D2T) systems~\cite{nlg_datatotext}. In the literature, it is possible to find text generation solutions for many different application domains, including health~\cite{nlg_stop1,nlg_neonatal}, environmental and weather information systems~\cite{nlg_TEMSIS,nlg_metofficedatatotext,bib_galiweather}, industry ~\cite{nlg_turbine}, project management~\cite{nlg_projectreporter} or education~\cite{nlg_dimitra,nlg_student2}. In recent times there has also been an explosion of commercially-applied D2T, due mainly to the increasing amounts of data that organizations have to handle\footnote{A short review of the most internationally recognized companies can be found in~\cite{bib_role_ldd_nlg}, but its number is likely to increase in the coming years}. Therefore, D2T systems are a fact in our society.


The main targets of D2T systems\footnote{In this paper we will refer mainly to D2T, but most of the statements made here about D2T also apply to NLG in general.} are human users and, therefore, the obtained texts must be, in addition to orthographically, grammatically and syntactically correct, also relevant, effective and persuasive. Consequently, choosing the best words and phrases that convey the most relevant information about the facts to be communicated deserves a special attention.


The generation of linguistic texts also needs to take into account the inherent features of natural language, such as vagueness~\cite{Russell1923}. The traditional approach to this issue from the perspective of D2T researchers is performed using numerical and symbolical crisp definitions supported by proper experiments~\cite{bib_acquiring_reiter,nlg_mousam}). In other words, in applied D2T systems, a vague word or expression such as `tall' or `in the morning' is usually defined by means of a crisp numeric interval such as $[175cm,300cm]$ or $[8:00am,11:30am]$ respectively. Consequently, we cannot say that vagueness is explicitly modelled but that a crisp definition is assigned to them; as a result, $174cm$ or $11:31am$ are not ``tall'' or ``in the morning''.\par

\begin{figure}[ht]
	\centering
	\includegraphics[width=0.8\columnwidth]{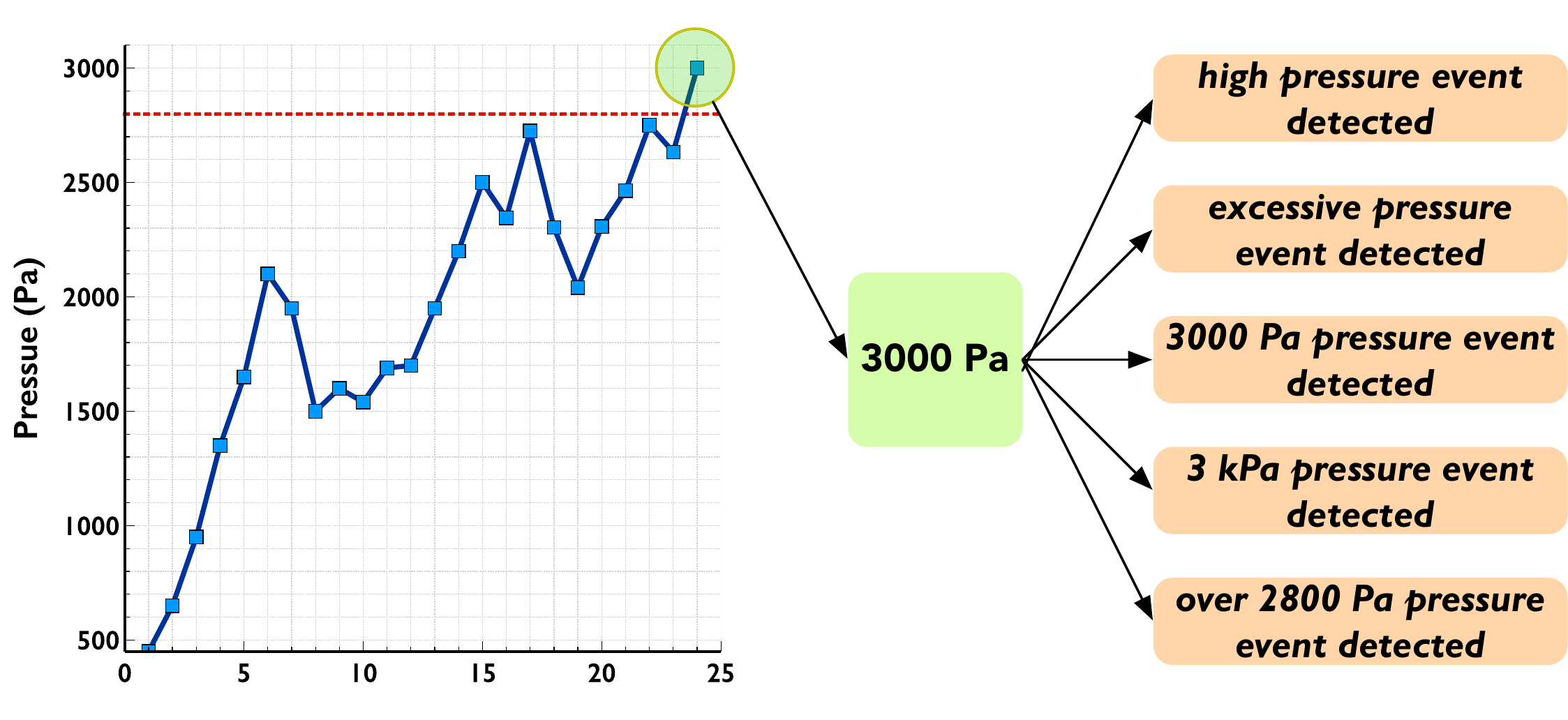}
	\caption{There are many ways in which the information that even a single numeric value holds can be expressed. This is a common problem that D2T systems have to face and leads us to consider the role that communicative intentions plays in human language.}
	\label{expr_sel}
\end{figure}

Vagueness is not only a matter of dealing with predicates involving borderline cases, such as the mentioned ``in the morning'' or ``tall'', but it also affects the degree of truthfulness or reliability of the statements. In these cases, we talk about uncertainty rather than vagueness, but they can be considered as two sides of the same coin. For instance, suppose an industrial process where an NLG system needs to report the evolution of the pressure in a valve (see Fig.~\ref{expr_sel}) during the last 25 minutes.

An immediate answer is to generate a set of twenty-five statements, e.g., \emph{in minute 1 the pressure was 490 Pa, in minute 2 the pressure was 650 Pa, in minute 3 the pressure was 950 Pa, etc.} This result is not useful for a human user, it needs to be rephrased preserving the same truth conditions. A good candidate is the sentence \emph{almost all the pressure values were under 2800 Pa}, where, introducing a vague quantifier such as `'`almost all'', the text can be significantly simplified without sacrificing its reliability.\par

\begin{figure}[ht]
	\centering
	\includegraphics[width=0.4\columnwidth]{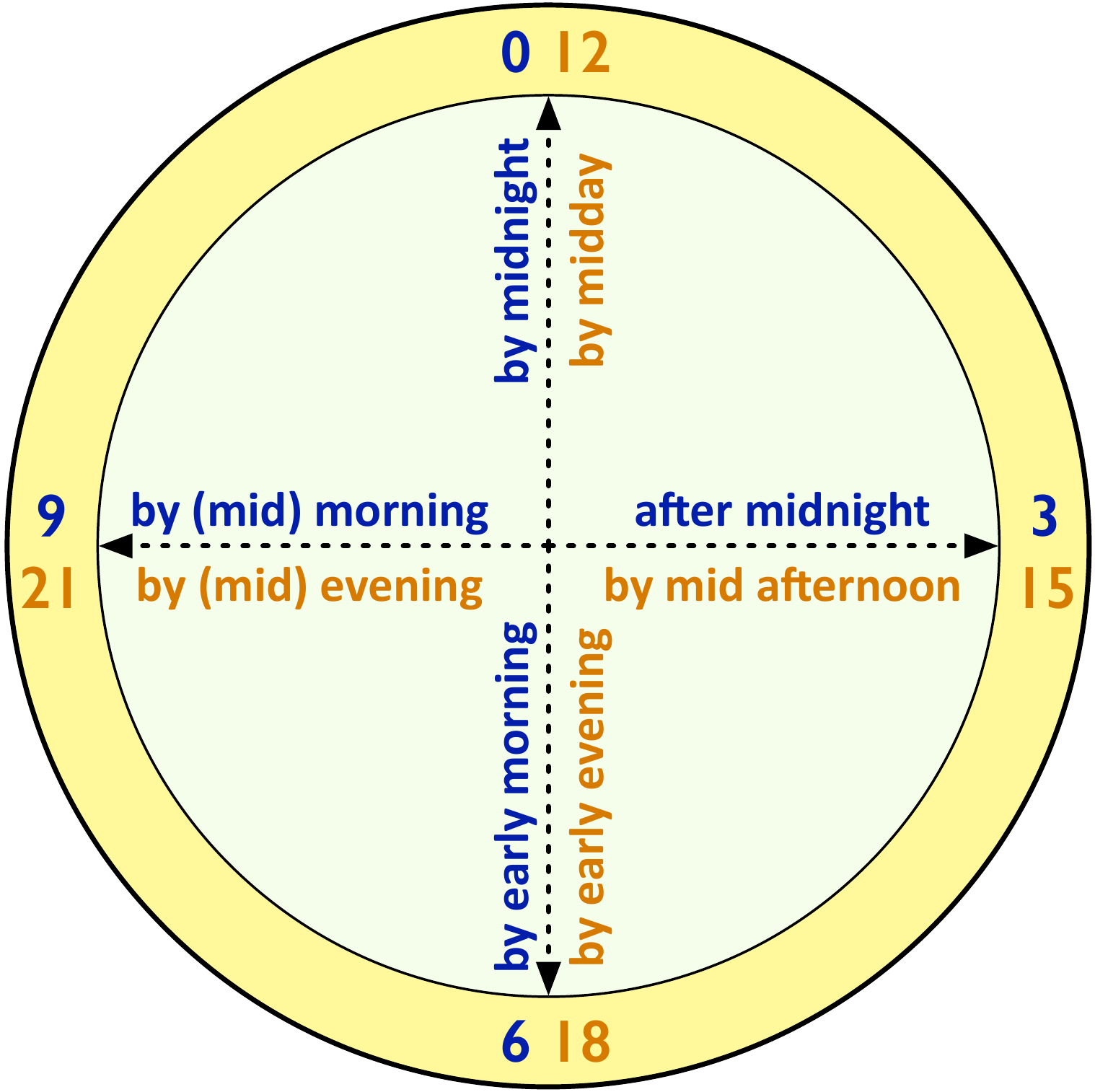}
	\caption{Definition of time phrases in the SumTime-Mousam system.}
	\label{mousamtimes}
\end{figure}

Although vagueness in D2T systems has not been explicitly addressed, this has not been an important impediment for obtaining excellent results~\cite{bib_acquiring_reiter,nlg_mousam} using alternative approaches. For instance, by means of the analysis of corpora (in order to capture how human writers or speakers actually use qualitative terms) and psycho-linguistic experiments, NLG researchers can comprehend how human users understand qualitative terms and define crisp definitions accordingly. The SumTime-Mousam system is a good example of this, where the authors used a corpus of 1045 manually written forecasts to analyse the correspondence between the use of time phrases and crisp numerical times (see Fig, \ref{mousamtimes}) appearing on the numerical datasets~\cite{nlg_mousam}. The evaluation of this system showed that readers preferred the automatically generated forecast texts over the manually written ones.

From our point of view, although these methods are effective, they are difficult to generalize because they are significantly time-consuming tasks. In addition, we disagree with the assumption that linguistic vagueness is only an epistemological issue that can be addressed searching for a crisp definition with enough information and time; in our opinion, vagueness is an ontological feature and should be analysed using different tools. A good example that illustrates the complexity of vague predicates is the \emph{Sorites paradox}~\cite{sorites}. Let us suppose that a person 185cm height is tall. If the person decreased in height by 0.5mm every night, when could it stop being considered tall? Such questions lead to the conclusion that vague concepts such as ``tall'' or ``heavy'' do not have a crisp meaning in certain contexts.\par


The use of fuzzy sets and fuzzy logic for the management of vagueness in the generation of linguistic texts from data was ignited by the research on fuzzy linguistic description and summarization of data (LDD)~\cite{bib_kacprzyk3}. LDD focuses on the extraction of imprecise linguistic information from numeric datasets~\cite{bib_role_ldd_nlg,bib_overview_methods_ldd}, but their impact and influence on D2T and NLG in general has been mostly residual, although some efforts have started to appear in this realm~\cite{bib_ldd_time_series,bib_role_ldd_nlg,bib_kacprzyk5,bib_galiweather,bib_gatt_portet_uncertain,bib_fuzzygre,bib_gradualrefexpr,bib_gradref2}. In spite of these efforts, we believe there is a lack of a proper in-depth discussion that justifies the use of fuzzy sets in D2T, and that involves the main nexus that relates one realm with the other: vagueness in human language.



We propose building a bridge between D2T and FST on the following question: \emph{Why should vague terms be tackled in D2T systems using FST?}; which can be split into four main sub-questions:

\begin{enumerate}
\item What does vagueness actually mean in FST? 
\item What does vagueness actually mean in D2T systems?
\item Why is FST a good theory for addressing vagueness in D2T systems?
\item Could D2T avoid FST for its improvement?
\end{enumerate}


For providing a clear answer to these questions, in Section~\ref{sc:vaguenessFST-NLG} we address the first two questions by means of a critical review of the literature about vagueness modelling in FST and NLG; in Section~\ref{sec:comp} we compare both paradigms in terms of their interpretation of vagueness and discuss why fuzzy sets could be a positive contribution for the development of D2T systems; in Section~\ref{sec:avoid} we describe under which circumstances FST can be dispensable. Finally, Section~\ref{sec:conclusions} highlights the main ideas discussed in this paper and provides a look at future trends about the use of FST in D2T.


\section{Understanding Vagueness in Fuzzy Sets Theory and D2T}
\label{sc:vaguenessFST-NLG}
Until the twentieth century, vagueness had a predominant negative consideration: vague predicates were defective ones due to lack of precision, and this could be easily solved by adding the missing information. Nevertheless, B. Russell, in his seminal paper Vagueness~\cite{Russell1923} in 1923, rejected this idea. He proposed that vague predicates are essential to natural language and allow us to denote those concepts which cannot be precisely defined, such as borderline cases. Good examples are the \emph{Sorites} paradox, which has been described in the introduction, or gradable adjectives, such as \emph{tall}, where the extreme cases are very clear (somebody who is 2 m. height is undoubtedly \textit{tall} and somebody who is 1,40 m. height is undoubtedly \textit{short}), but the ones that are in-between constitute a penumbra area where the change from one extreme to the other is fuzzy.\par

As a result of this new conception of vagueness, classical conceptions of truth and falsehood based on the excluded middle law must be reconsidered. Thus, statements are not true or false anymore but they should be qualified by means of a degree of trustworthiness (e.g., \L{}ukasiewicz's logic~\cite{Lukasiewicz1970} introduces the intermediate value of ``possible''), Bayesianism appeals to agent's degree of belief for these intermediate values~\cite{Adams1998}, etc.).


In this section, we will analyse in detail the perspective to tackle vagueness proposed by fuzzy logic, widely adopted in Computer Science, and the one adopted by NLG, which underlies the majority of systems in this discipline used in real applications.

\subsection{Vagueness from a Fuzzy Sets Theory Perspective} \label{sec:vague_fst}
The notion of a fuzzy set was proposed by Zadeh~\cite{bib_Zadeh2} and formalizes the insight of linguistic vagueness in terms of borderline cases by means of the concept of gradualness in class membership~\cite{Dubois1997}; using Zadeh's words, ``a class of objects with a continuum grades of membership''~\cite{Zadeh1965}. This formalization is known as membership function $\mu$ and it is defined in the interval $[0,1]$.\par

Notwithstanding, the meaning of a membership grade is a debatable question in FST and there is not uniformity about it. During the last fifty years, FST has been developing both from a mathematical point of view~\cite{Cintula2015} and an engineering one~\cite{Azar2014} and several different semantics for the notion of degree of membership have been proposed. In this section, we will analyse the three semantics for fuzzy sets (i.e., similarity, preference and uncertainty) described by D. Dubois and H. Prade~\cite{bib_Dubois1997} and the semantics proposed by the paradigm of Computing with Words (CWW)~\cite{bib_Zadeh}. 

\subsubsection{Three semantics for fuzzy sets}
Fuzzy sets seem to be applied in three main different basic problems~\cite{Dubois1997}: classification and clustering, decision-making problems, and approximate reasoning. Although all of them use the concept of degree of membership in the same form $\mu_{F}(U)$ (an element $u$ belongs to a fuzzy set $F$ defined in a referential $U$ with a degree in the interval $[0,1]$), there are different alternative underlying interpretations.\par

Classification and clustering problems usually interpret the membership function in terms of similarity, because the elements are classified according to their inherent features. Thus, given a fully representative element (named prototype), which $\mu_{F}(u)=1$, all the remaining elements of the universe are sorted according to their resemblance to the prototype in terms of a distance function, which generates the corresponding values for a $\mu$ function; those elements that are totally different to the prototype gets $\mu_{F}(u)=0$. A very well-known example of this type of semantics is the  clustering task using The Iris Data\footnote{Iris Species Database httop://www.badbear.com/signa/
}, where a collection of flowers are classified into three fuzzy classes (\emph{Iris setosa}, \emph{Iris versicolor} and \emph{Iris virginica}), according to different measurements (sepal length/width and petal length/width). The vagueness of this task relies on the categories, not on the measurements, because there is not a sharp border that distinguishes one category from the others.


In decision-making problems, on the other hand, fuzzy sets are devoted to the modeling of flexible criteria or constraints rather than resemblance features. Thus, a fuzzy set is, in general, a collection of values of a decision variable $x$, where the membership function $\mu_{F}(u)$ indicates the degree of preference of the user for the value $u$. The most preferable a value is, the higher its degree of membership to the fuzzy set. An example of decision-making problem is to choose a good car according to the following criteria, which involves fuzzy terms:
\begin{enumerate}
\item It should have a good average of litres of combustible per km.
\item It must be safe.
\item It must be cheap.
\end{enumerate}

Each one of these three criteria is defined by means of a fuzzy set; e.g., $average\ l/km = {0.8/good, 0.5/medium, 0.2/bad}$, where the values of the variable $average\ l/km$ are sorted according to the preferences of the user expressed in the criterion 1. In this case, this variable is not so relevant (it is qualified by the modal verb \emph{should}), but the preference in favour of the \emph{good} value is clear with respect to the other two (\emph{medium, bad}).

The last interpretation corresponds with typical cases of approximate reasoning, it is usually named as the possibility theory~\cite{Zadeh1978}. Thus, a fuzzy set is a set of possible values or parameters ($U$) of a variable $x$ and the membership function indicates the degree of possibility of one of the parameters happening. In this case, it is known that $x$ takes one of the values of $F$ and the degree of membership indicates the degree of belief of an agent o which particular value $u$ is taken by the variable. In~\cite{Amgoud2004}, we can find an example of applying possibility theory to a dialogue game in deliberative negotiations. Here, possibilistic logic is used both for representing the mental states of the agents involved in the dialogue, but also for revising the bases and describing the decision procedure. For instance, the possible movements of an agent are modelled as a fuzzy set and the degree of membership indicates how possible a move in the process is.

\begin{table}[thb]
\centering
	\begin{tabular}{|l|c|c|c|}
    \hline
					&Similarity	&Preference		&Uncertainty\\\hline
      Elements		&objects	&values			&values/objects\\\hline
      Perspective	&objective	&intentional	&subjective\\\hline
      Structure		&prototype	&non-prototype	&non-prototype\\\hline
      Measurement	&distance	&cost			&frequency\\
      \hline
	\end{tabular}
    \caption{\label{tab:SPU} Comparison among the three semantics for fuzzy sets.}
\end{table}

In Table~\ref{tab:SPU} we summarize a comparison among the three semantics according to different features. Similarity semantics usually deals with collection of physical objects, which can be precisely measured, but the belonging categories are very difficult to define sharply. A significant difference of this proposal with respect to the other ones is the notion of prototype, because it guarantees an element with the maximum degree of membership in the fuzzy set. Preference semantics, on the other hand, relies on fuzzy sets whose elements are values of a variable, and not physical objects, which convey the preferences of the user with respect to a particular decision. Given that, most of the times, decision-making has to satisfy multicriteria, the best way for assessing the degree of membership is in terms of cost for achieving the desired goal. Finally, uncertainty semantics, is the most subjective one, since it captures the user's belief degree with respect to the possible values of a variable.

\subsubsection{Fuzzy sets in Computing with Words} \label{cww}
In the mid-1990s, Zadeh introduces the paradigm of computing with words (CWW)~\cite{bib_Zadeh4}, a new way of computing where the computational operations are executed by means of words instead of numbers. As a result, natural language, both from a semantic and syntactic point of view, becomes a key tool for human-machine interaction, mainly in problems that involve too much imprecision to be solved in the traditional numerical way.\par

Under this new paradigm, the concepts of \textit{granule} and \textit{protoform} appear~\cite{bib_Zadeh}. A granule is defined as ``a clump of physical or mental objects (points) drawn together by indistinguishability, similarity, proximity of functionality''~\cite{bib_Zadeh2}. Each granule, which can be crisp or fuzzy, is the basic processing unit of information and there are four criteria that guide its definition~\cite{bib_Zadeh4}:
\begin{enumerate}
\item It must be small enough to be manipulable.
\item It must provide relevant insight about the problem in order to make it understandable.
\item Its origin must be objective numerical values.
\item Each granule must represent a relevant part of the problem in order to be addressed.
\end{enumerate}

For instance, let us consider the temperature in a meteorological service. The temperature captured by a 
thermometer is registered in a table every 30 minutes. A case of precise granule is to use the Celsius scale considering only degrees and half degrees, therefore, two temperatures such as $25.7º$ and $25.8º$ are indistinguishable in our register because of both will be represented as $26º$. A case of a fuzzy granule is to generate a fuzzy partition of temperature ranges using the following labels: \emph{very cold}, \emph{cold}, \emph{warm}, \emph{hot}, \emph{very hot}; in this case, with \emph{warm} defined as a trapezoidal function $\widetilde{warm}={20,22,24,26}$, temperatures such as $21º$ and $25º$ will be registered in our system simply as \emph{warm}, with the corresponding degree of membership.\par

The concept of granule is usually expressed by a proposition $p$ with the following form,
\begin{equation}
	X\ isr\ R
\label{eq:granule}
\end{equation}

\noindent
where $X$ is a constrained variable, $R$ are the information granules and $r$ denotes the type of relationship between $X$ and $R$. In~\cite{bib_kacprzyk3}, 10 different $r$ are defined; for instance, in the example of the fuzzy granule temperatures, $r$ is (fuzzy) disjunctive because of a temperature belongs to one label or adjacent ones, i.e., it is not the case that a temperature is \emph{very cold} and \emph{very hot}.\par

The last step are the encoding and decoding mechanisms, which capture the objective data according to the defined granules. For instance, in the example of the temperature again, it is necessary to define a membership function for mapping the temperatures of the thermometer to the corresponding fuzzy labels.\par

The second notion to be considered is the \emph{protoform}, which is related with the output mechanism from the granule. A protoform is defined as an abstract prototype of a linguistic summary~\cite{bib_kacprzyk3},

\begin{eqnarray}
Q\ Y's\ are\ S\\
Q\ KY's\ are\ S
\label{eq:prototype}
\end{eqnarray}

\noindent
where $Y$ is a set of objects, $K$ is a qualifier, and $S$ is a summariser. The concept of summariser $S$ is directly related with the communicative intention, and it conveys the set of attributes to be predicated. In addition, it also introduces the linguistic quantifier operator $Q$, which is a flexible aggregation operation. For instance, if $Y$ denotes a set of students and $S$ the set of possible marked qualifications, ``few students have obtained a good mark'' refers to the subset of students with \emph{good marks} but also is a way of summarising or describing the information about the grading of the whole set of students. Likewise, if $K$ refers to the gender of the students, we can also provide a description such as ``few male students have obtained a good mark''. In this context, the concept of protoform is linked to the concept of fuzzy quantified sentence (type-I in the first case, and type-II where the qualifier $K$ appears). In fact, both protoforms and fuzzy quantified statements are often used interchangeably to refer to the same idea.\par


An actual application of these concepts is the realm of ``linguistic summarization or description of data'' (LDD)~\cite{bib_Yager}. An LDD system extracts, by means of granules, relevant information from numerical data and generate short linguistic excerpts. There is an extensive collection of research work in this topic (see the following reviews of methods and use cases for further information~\cite{bib_overview_methods_ldd,bib_role_ldd_nlg,bib_ldd_time_series}) and in the recent years there has been an increasing effort in converting those short linguistic pieces into textual phrases useful for end-users. 

\subsection{Vagueness from a Data-To-Text Perspective} \label{sec:vague_d2t}

There is an essential difference between FST and NLG about how vagueness is tackled: while for FST it is a matter of an accurate representation of imprecise information, for NLG it is a matter of efficiency to achieve the communicative goal of the speaker according to the context. Thus, for the latter, the use of vague predicates or expressions is part of the strategy to select the most adequate wording in order to achieve a predefined goal.\par

Despite of this, as in FST, it is easy to identify the two aforementioned dimensions of vagueness: vague predicates for referring borderline cases (e.g., ``X is tall'', ``X is short'', etc.) and uncertainty for assessing the reliability of assertions (e.g., "X may have happened'', etc.). In this section, we will analyse the different approaches developed in NLG for handling both dimensions of the same problem.


\subsubsection{Borderline Predicates in NLG}
In the field of NLG, we can find two main viewpoints, mainly opposed, about vagueness. The first one, substantially supported by K. van Deemter, claims that vagueness should be generally avoided because it is a source of ambiguities and misunderstandings and, as a consequence, handling vagueness is not a core part in the NLG systems. On the other hand, as A. Gatt et al. hold, NLG systems must represent adequately vagueness, because it is an inherent characteristic of natural language and it cannot be avoided in many cases of referring expressions.


Kees Van Deemter, in several papers~\cite{nlg_vandeemter06,vanDeemter2009,nlg_vandeemter09}, argues that Game Theory is a good theoretical framework for analysing the utility of vague expressions in NLG systems. In particular, he focuses on gradable properties in referring expression generation\footnote{A referring expression, following its typical definition in linguistics, is any noun or phrase, or surrogate for a noun phrase, whose function in discourse is to identify some individual entity (such as an object or a person).}. Thus, the use or lack of use of a vague expression is determined according to the context and the strategy to achieve the goal of the speaker.\par

A good example that illustrates this idea is described in~\cite{kees2000generatevague}, which is a modification of the referring expression generation algorithm proposed by Dale and Reiter~\cite{nlg_griceandale}. Let us suppose a domain of five mice sized as $3,4,5,10,11$ cm and the expressions ``the largest mice'' and the ``the largest 2 mice''. The former is clearly more vague than the latter, because of removing the numeral entails a loss of information. Therefore, in order to avoid ambiguity and misunderstandings, the second expression is preferable than the first one, although this is the shortest one. However, there are two circumstances where the former is more adequate: (i) when any ambiguity resulting from the different values of the numeral is not relevant; (ii) when 'natural grouping'\footnote{Intuitively, the difference between two adjacent members in the scale is comparatively small; e.g., the mice with 10cm and 11cm are a natural group given the difference with the third one, 5cm, which is much bigger} is allowed by the domain.\par

Another possible use of gradable adjectives is the selection of their form (base, comparative and superlative) according to the context and the communicative aim of the speaker~\cite{kees2004vaguetuning}. This question is addressed from an experimental point of view, using corpora studies and pilot experiments, and the conclusion reached is that base forms might be preferred over the superlative ones. As in the previous study, some exceptions can appear, such as the subjective preferences of the speaker, but an analogous position is supported: crisp predicates are preferable than vague ones in referring expressions.

A third part in the analysis of vague referring expressions can be found in their use as mechanisms to present data into a human-accessible form and suppress irrelevant details (losing the irrelevant information). Initially, this seems to be an adequate use for them but several important issues arise: (i) the complexity of determining which is the best expressive choice given a particular context increases when more possible options are available (see Fig.~\ref{expr_sel}); or, (ii) the so-called multidimensionality issues, where more than one vague adjective must be considered by the algorithm for generating the adapting referring expression.

As a result of these different analyses conducted by Van Deemter, it can be inferred that, in general, vague expressions should be avoided in order to generate the most clear and understandable texts. Although they might be useful in some specific contexts, a crisp wording is more efficient and effective from a communicative point of view when precise data are available.\par   

Despite this conclusion, van Deemter himself recognizes that there are still some open questions about the role of vagueness from a communicative point of view. In~\cite{nlg_vandeemter09}, he explores two main questions: why vagueness appears in language and when and why a speaker should choose a vague expression rather than a precise one. In his analysis, he concluded that there are some circumstances where vague predicates might appear, such as when terms are essentially vague (e.g., ``cloudy''), there exists a cost reduction (vague expressions are easier to produce and interpret than crisp ones), future contingencies (as in weather forecasting) or lack of good metrics (if the system cannot provide accurate crisp expressions, it might use vague ones instead) and, therefore, the use of vague expressions is a matter of choice. However, this does not invalidate his previous conclusions from other studies and van Deemter concludes, again, that vague expressions should be avoided generally.

Other authors that support this same approach to vagueness are Power and Williams~\cite{nlg_power12}, which propose the use of numerical approximations to describe proportions at different levels of precision. Thus, if we compare phrases such as ``25.9 per cent'' and ``more than a quarter'', the latter is more vague than the former, because of there exists a loss of information with respect to the crisp value.\par

A. Gatt et al., on the other hand, provide a different perspective about the impact of vagueness in NLG systems~\cite{bib_gradualrefexpr}. They explore the case of referring expression generation from non-linguistic data, where the use of fuzzy terms such as colour or position in an image, even being fuzzy concepts, can be more useful than crisp expressions, since these are very difficult to be sharply defined. Their conclusions are supported by an experiment using an image of labelled human cells, where different examples of referring phrases were compared according its referential success degrees. The achieved results support the claim that vagueness is an issue that cannot be generally avoided in NLG.\par

Other example of real application where vague concepts play an important role is GALiWeather~\cite{bib_galiweather}, a D2T system which generates textual weather forecasts from short-term prediction data. Concepts such us ``beginning'', ``predominant'', etc. are modelled by means of fuzzy sets given the impossibility of getting a crisp definition for them.



For concluding, the case of borderline predicates in NLG is mainly dealt with in the area of generating referring expressions. In the literature, we can find two main opposed points of view; one of them argues that vague or fuzzy expressions should be avoid whenever possible because they entail a lack of information and communicative efficiency. The other one holds that there are non-linguistic data whose verbalization is inherently vague, therefore, specific theoretical tools for modelling vagueness, such as fuzzy logic, must be used in order to preserve the adequate degree of representativeness of the linguistic description.


\subsubsection{Uncertainty and NLG}
\label{ssc:UncertaintyNLG}
Uncertainty is another dimension of vagueness to be considered in NLG systems. In Reiter's words~\cite{nlg_datatotext}, \textit{applied NLG systems may need to communicate uncertainty about the reliability of the input data or the system's analysis}. The main contribution to this area are the recent works of Gatt and Portet, whose main target is to tackle uncertainty in the modeling and conveyance of temporal expressions~\cite{nlg_impreciseportet,bib_gatt_portet_uncertain}.

In~\cite{nlg_impreciseportet}, Porter and Gatt introduce fuzzy techniques to deal with uncertainty in temporal data series used in the \textsc{BabyTalk} family of systems. These generate reports from a signal analysis about the clinical state of babies that are in neonatal ICU. Handling these temporal events generates uncertainty in the production of linguistic statements and they apply the fuzzy theory of possibility~\cite{dubois2015possibility} in order to choose the most adequate modal expression for them (e.g., may, must, etc.); for instance, \textit{The baby was moved from SIMV to CPAP. He was extubated and underwent oral suction. This must have caused the instability in HR and SpO2}.




In~\cite{bib_gatt_portet_uncertain}, the same authors advanced a step further, with a theoretical model based on FST which is combined with experimental data (from three different languages, French, Maltese and English) in order to capture the subjective bias of this kind of judgements involving uncertainty. As a result, they developed a classifier that selects the most appropriate uncertain expression according to the obtained possibility-necessity values and the subjective bias in order to enhance the feasibility of using the model as an underlying mechanism for an NLG system.\par

For concluding, we can say that vagueness has not been a core issue in the development of NLG systems. In addition, until recently, the most dominant claim in the literature was that there were not strong empirical or pragmatic reasons to improve the representation of borderline predicates or uncertainty; it was enough to select a crisp definition based on data to assign a meaning to them. However, in the last years, a new perspective has arisen, arguing that vagueness is relevant and needs to be specifically tackled.

\section{Comparing and Integrating D2T and Fuzzy Sets in Terms of Vagueness Interpretation} \label{sec:comp}

From our perspective, the current lack of understanding between D2T and FST relies on their different perspectives on vagueness; while the former adopts a pragmatism conception of this sort of expressions subordinated to its communicative function, the latter focuses on enhancing the accuracy of the mathematical representation.

Nevertheless, as the studies by Gatt et al. suggest, the current state of the art in D2T technology seems to demand an specific treatment of vagueness. In order to bridge this gap, it is essential to connect both perspectives and explain the benefits that FST can bring to D2T/NLG in a proper way, i.e., providing an answer to the third question listed in Sec. \ref{sec:intro}.

A contribution to this mutual understanding can be found in~\cite{bib_kacprzyk3}. In this paper, Kacprzyk and Zadro\.zny identified in a potential connection between fuzzy sets and NLG through fuzzy linguistic summarization, and discussed that NLG could benefit from the imprecision or vagueness treatment that FST offers. Likewise, they argued that NLG techniques could be used to provide a text generation interface for linguistic summarization. These ideas were later retaken in \cite{bib_kacprzyk5} without noticeable advances.

\subsection{Why is FST a good theory for addressing vagueness in D2T systems?} \label{good}
At this point, we will further develop the ideas described in \cite{bib_kacprzyk3} and \cite{bib_kacprzyk5}. We believe that a more in-depth analysis about the underlying semantics of fuzzy protoforms will help shed light on the benefits that FST and LDD can bring to D2T. For this, we will describe a simple example based on a type-I fuzzy quantified statement that merges both FST and D2T perspectives.

Let us refer to the fuzzy protoforms described in Sec. \ref{cww} that adopt the form of \textit{Q Y's are S} (namely, type-I fuzzy quantified statements). Suppose, then, a D2T system tasked with generating descriptions about the pressure of a set of valves, $V=\{v_1, \ldots, v_n\}$ in an industrial environment. The system computes type-I fuzzy quantified sentences on a fuzzy linguistic variable that models the pressure, $P=\{low, medium, high\}$, using a set of fuzzy quantifiers $FQ=\{few, several, many\}$. Suppose also that we are certain that both $P$ and $R$ represent properly the semantics of the linguistic terms according to the expert's knowledge.

Consequently, the system produces descriptions such as ``A few valves have high pressure'' or ``Most valves have low pressure'', which are later properly verbalized to match the domain language requirements. In this context, let us refer now to how a fuzzy quantified sentence is computed. There are several fuzzy quantification models \cite{bib_quantstate} that allow to calculate the truth degree of a quantified sentence, but for illustration purposes we will refer to Zadeh's model~\cite{Zadeh1985}, which is also the most widely used in the literature:

\begin{equation}
T(Q\ Y's\ are\ S) = \mu_Q\bigg(\frac{1}{n}\sum_{i=1}^{n}\mu_P(v_i)\bigg)
\label{type1}
\end{equation}

Suppose then, that we have the latest pressure measurement for our set of valves, and the D2T system generates a description of this situation. Using Zadeh's model, we calculate the truth degree of all possible combinations for $P$ and $R$. Translating the procedure in Equation \ref{type1} to words implies to evaluate each pressure measurement against each pressure fuzzy label, determine the fuzzy cardinality of the set of valves, and then evaluate this result against all fuzzy quantifiers.

After computing all possibilities, suppose that our system determines that $T(most\ valves\ have\ high\ pressure)=0.8$. What does $T$ actually mean and what implications does this value have for the D2T system, so that it can convey such information? If we backtrack the process, this starts from fuzzy definitions that represent vague terms about the pressure level, such as `low' or `high'. These definitions allow us to know, for each of the labels in $P$, the truth degree for ``$v_i$ is $P_j$'', i.e., $\mu_{P_j}(v_i)$. Given that $P$ was defined to reflect the knowledge of an expert, the truth degrees resulting from this evaluation are easy to interpret, as they are a good representation of the degree in which each pressure value is `low',`medium', and `high' according to the expert.

However, by performing the remaining quantification procedure displayed in Eq. \ref{type1}, which involves obtaining the fuzzy cardinality for the whole set of valves, and then evaluating against a fuzzy quantifier, we end up aggregating the original individual truth degrees into a single truth degree that represents something far more complex than the correspondence degree between a pressure value and a vague term. One direct interpretation of $T(most\ valves\ have\ high\ pressure)$, is ``the degree in which the `high' pressure values fulfil that they are `most' within all values''. This corresponds to a rather logical interpretation that follows from the mathematical formula in Equation \ref{type1}. In our case, however, we are interested in an interpretation closer to a language use perspective, which can be useful for text generation purposes.

Considering that we are dealing with vague terms defined as fuzzy sets, such as `low' and `most', another complementary interpretation of $T$ that can be useful for deciding how to convey quantified sentences is that the imprecision of these vague terms ends up causing a lack of certainty in what has to be stated. For instance, in our example, $T(most\ valves\ have\ high\ pressure)=0.8$ would mean, according to this interpretation, that we are rather certain (or, alternatively, that there exists an important evidence) that ``most values have high pressure''. This interpretation provides several benefits for our D2T system:

\begin{itemize}
\item The D2T system may choose to rank and/or discard statements according to their level of certainty.
\item The D2T system can decide to verbalize the statements either alone or complemented by an assertion that communicates the level of certainty about each statement.
\item The D2T system may communicate situations of ambiguity or conflict, when two or more statements share a similar level of certainty.
\end{itemize}

This understanding of fuzzy quantified statements is aligned with Reiter's postulates about D2T systems in~\cite{nlg_datatotext}, where, as stated in Section~\ref{ssc:UncertaintyNLG}, these \textit{may need to communicate uncertainty about the reliability of the input data or the system's analysis}. This idea is not only applicable to the LDD framework, but also to computing with words and FST in general. Under this new light, FST provides a framework that allows to model the vagueness of terms and expressions to be used, as well as to manage the uncertainty that results from data analysis.

Certainly, we have also seen that LDD is not the only connection of FST with D2T and NLG. In Section~\ref{sec:vague_d2t} we have reviewed the uncertainty interpretation of fuzzy sets which was used by Portet and Gatt in~\cite{nlg_impreciseportet} to model and convey uncertain temporal expressions through possibility theory. Likewise, the idea of considering vague (fuzzy) properties in the task of referring expression that was proposed by Gatt et al. in \cite{bib_gradualrefexpr} and the use of fuzzy temporal labels in \cite{bib_galiweather} responds to the similarity interpretation of fuzzy sets (e.g., if we have a fuzzy definition of \textit{small}, we can calculate the degree in which objects or values match our prototype of \textit{smallness}).

Other interesting D2T problems that can be addressed using fuzzy sets, and specifically LDD techniques, include generating temporal and geographical referring expressions. Time series data have been a recurring resource and research interest for LDD since its inception \cite{bib_ldd_time_series,bib_overview_methods_ldd}, and its application in D2T is. However, the most interesting problem from a D2T perspective is the generation of geographical referring expressions, which was studied and applied in the RoadSafe system \cite{nlg_roadsafe2,nlg_roadsafe}. In short, it is the problem of determining the best expression that refers to a set of geographical points where a relevant event (which will be described within the automatically generated text) takes place. The treatment of geographical descriptors, such as `North', `coastal' or `inland' that RoadSafe employed for the generation of geographical references was based on a crisp grid approach that did not consider their inherent vagueness with respect to the use that the experts made of such terms. In this context, FST and LDD will allow to improve, as the preliminary study in \cite{bib_fuzzygre} shows.

\subsection{Moving towards the development of applied D2T systems}

Although merging the understanding of vagueness and uncertainty under the same umbrella is a necessary step to allow FST and D2T to unite from a theoretical and discussion perspective, in order to extend this interpretation to actual systems, a proper methodology that merges standard practices from both disciplines is needed, as proposed in \cite{bib_fuzzygre}. The discussion corresponding to this issue is out of the scope of this paper, but given its importance we will provide a brief commentary and a few examples about this topic.

\begin{figure}[h]
	\centering
	\includegraphics[width=0.7\columnwidth]{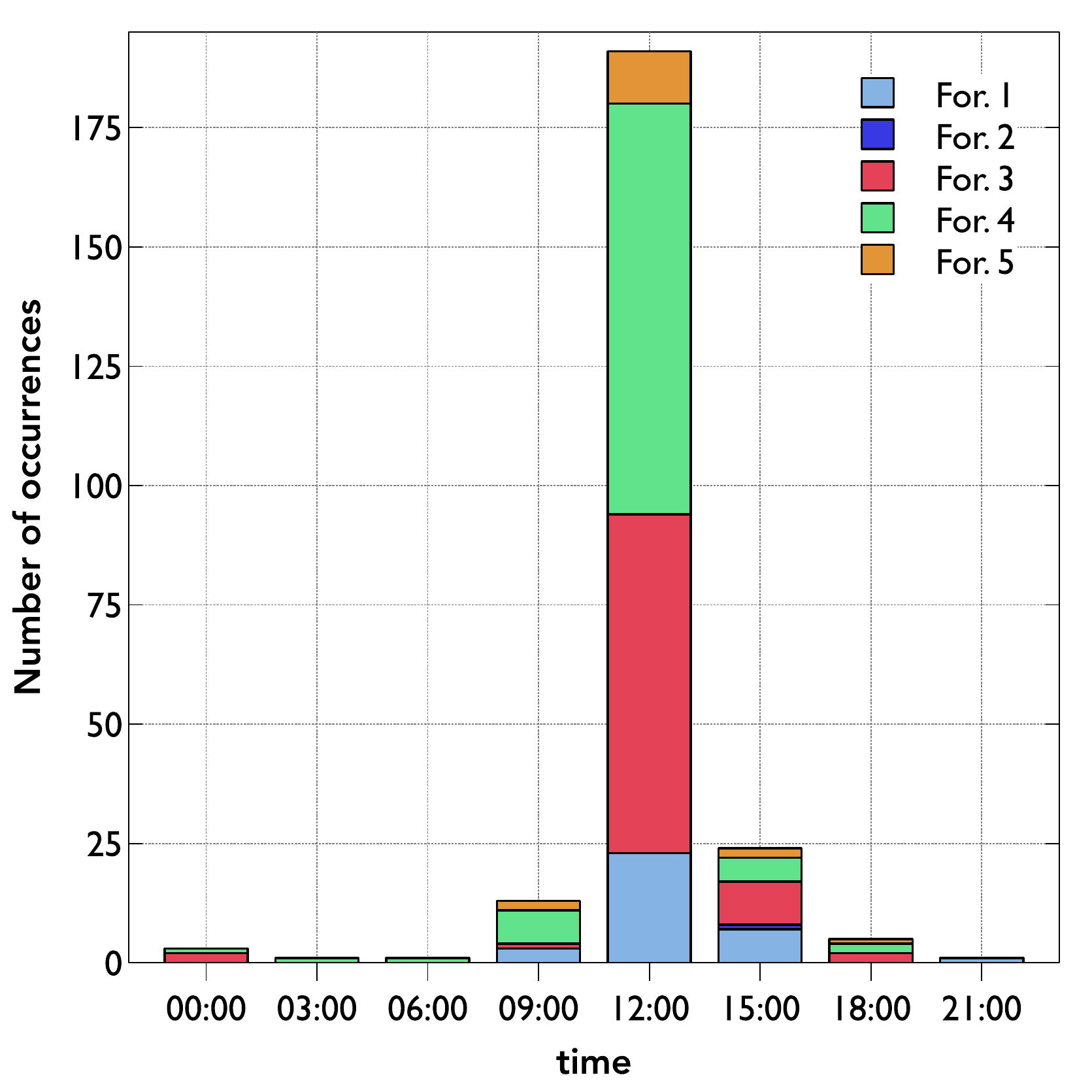}
	\caption{Histogram that relates the actual usage of the ``midday'' expression by five different forecasters in the corpus of the SumTime-Mousam project.}
	\label{midday}
\end{figure}

In short, the application of FST in D2T systems will require to adapt existing experimentation approaches in D2T to model vague terms and expressions as fuzzy sets, as well as to study which formalisms from FST and LDD can be applied to the obtained models \cite{bib_fuzzygre}. For instance, the temporal expressions used in the SumTime-Mousam system \cite{nlg_mousam} that resulted from the analysis of the corpus forecasts could have been modelled as fuzzy sets to represent the different but overlapping interpretations of the five forecasters that authored the corpus texts. In Fig. \ref{midday}, it is shown that the expression ``midday'' could be gradually defined according to its actual usage.

\begin{figure}[h]
	\centering
	\includegraphics[width=0.7\columnwidth]{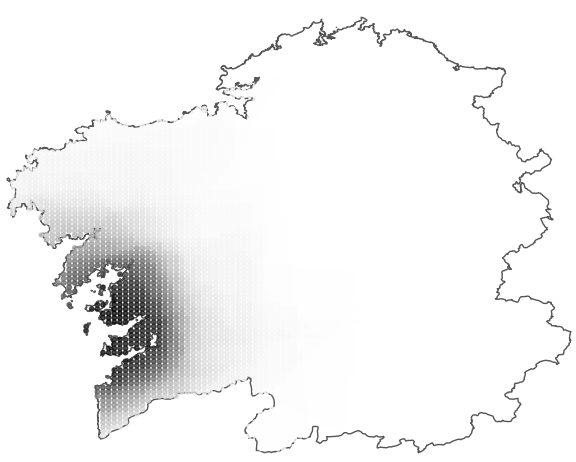}
	\caption{Representation of the fuzzy geographical descriptor ``R\'ias Baixas'', located within the Galician region in Spain.}
	\label{riasbaixas}
\end{figure}

Likewise, in the realm of geographical referring expression generation, to empirically determine the understanding of geographical descriptors by experts or readers, such as ``North'' or ``coastal'' (which are inherently vague and gradual by nature \cite{nlg_roadsafe2}), is essential to build a proper D2T system. This requires adapting typical data acquisition techniques from users such as surveys, and to use such data to build fuzzy representations of the geographical concepts, as was done in \cite{model_fuzzygeo}. For instance, Fig. \ref{riasbaixas} shows a fuzzy geographical descriptor  that represents the region ``R\'ias Baixas''.

Thus, to create empirical fuzzy models of linguistic terms and expressions, and to study which mechanisms from FST can be appropriate for D2T systems and related problems, are some of main the challenges ahead that need to be explored in order to help move forward the integration of FST into D2T.
\section{Could D2T avoid FST?} \label{sec:avoid}
As a promising framework, LDD and FST in general are tools that can improve the modelling and managing of vagueness and uncertainty in D2T. Among the benefits that can be accounted for, FST allows to create a more accurate representation of the actual usage of words and expressions by human users, and can also be used to communicate within the generated texts in an implicit or explicit way information about their (lack of) certainty. 

This does not necessarily mean, however, that when a D2T has to generate texts that include vague words, these have to be modelled by means of FST. In fact, there can exist several situations where the application of FST in D2T would not be feasible or advisable, for instance,
\begin{itemize}
\item during the knowledge acquisition stage of the development of the D2T system,
\begin{itemize}
\item if the empirical meanings of vague words or expressions to be utilised are given by expert guidelines that assign exact crisp numeric values or intervals to the terms (e.g. `cold'=[0,15]º).
\item if, after empirically studying the usage of words by human writers or readers (through psycho-linguistic experiments, surveys, or corpus analysis), these can be represented using crisp definitions without excessive loss of information.
\end{itemize}
\item because, due to the application domain of the D2T system, managing vagueness and uncertainty in the generated texts is not a priority.
\item because using a crisp approach simplifies the definition and management of words or expressions and is an acceptable trade-off for the system's performance, even if managing vagueness or uncertainty could be applicable.
\end{itemize}

To sum up, in addition to determine if textual information should be conveyed using vague expressions, one has to decide for each specific case if FST is a tool that fits the domain application of the D2T system and its requirements. For instance, FST could be avoided under the circumstances that we have listed above. Thus, answering in a general sense if D2T could avoid FST for its improvement should not be the right question (in fact, NLG has been developing for more than 30 years without resorting to FST).

Thus, this question, in our opinion, should be rephrased as whether or not D2T \textit{should} avoid FST to solve the problem of vagueness and uncertainty in this field. Based on our previous discussion in Sec. \ref{good} about the benefits that FST can bring to D2T, we strongly believe that FST should become the main framework that should be studied and applied for this task. Omitting FST at this stage of cross-fertilization between both fields would be, in our opinion, a missed opportunity at the very least.

\section{Conclusions} \label{sec:conclusions}
Vagueness, alongside uncertainty, are both important issues affecting D2T and NLG. However, these have not been treated extensively and remain as open challenges. In this sense, there exists an important potential use of FST for managing vagueness and uncertainty in D2T. Particularly, in this paper we have brought together the interpretations of vagueness and uncertainty from both disciplines and provided a more unified understanding for the integration of FST in D2T.

It is also interesting to point out that the first actual connections between D2T and FST that Kacprzyk and Zadr\.ozny discussed have not been established (from a genuinely NLG perspective) from the use of LDD approaches, but from the use of possibility theory and fuzzy constraint temporal networks in the works of Gatt and Portet, i.e., from an uncertainty interpretation of fuzzy sets. This does not invalidate at all the original idea of Kacprzyk and Zadr\.ozny, but rather seems to indicate that, as we have reviewed, there are at least two different complementary uses of FST in NLG.

Given the wide variety of problems that NLG offers, we expect to see in the coming years a better integration of FST techniques to address vagueness in D2T at many different levels. Current and future research trends in this regard include:

\begin{itemize}
\item Integration of FST-based techniques into NLG (such as fuzzy neural networks, genetic fuzzy systems or fuzzy rule-based systems).
\item Referring expression generation using fuzzy properties.
\item Modeling and conveying uncertainty using possibility theory or probabilistic logic. 
\item Construction of fuzzy models of linguistic terms and expressions based on data from experiments or corpus studies.
\item Application of fuzzy linguistic summarization/description of data techniques for content selection purposes.
\item Research on the influence of using FST on NLG tasks.
\end{itemize}

In the long term we expect that FST can be considered the \textit{de facto} framework for treating vagueness and uncertainty in NLG. Thanks to this new theoretical underpinning, the most difficult and harder tasks that need to be done in the early stages of NLG systems, such as the extensive experiments or corpus studies, will be less exhaustive, resulting in a faster development of NLG systems. Likewise, we expect that the construction of models of vague expressions based on such techniques will allow to reflect human language use more faithfully and to design algorithms which perform more effectively the selection and conveyance of such expressions, thus improving the overall performance of D2T systems in terms of communication success.

\section*{Acknowledgements}
This work has been funded by TIN2014-56633-C3-1-R and TIN2014-56633-C3-3-R projects from the Spanish ``Ministerio de Econom\'{i}a y Competitividad'' and by the ``Conseller\'{i}a de Cultura, Educaci\'{o}n e Ordenaci\'{o}n Universitaria'' (accreditation 2016-2019, ED431G/08) and the European Regional Development Fund (ERDF).  A. Ramos-Soto is funded by the ``Conseller\'{i}a de Cultura, Educaci\'{o}n e Ordenaci\'{o}n Universitaria'' (under the Postdoctoral Fellowship accreditation \textbf{481B 2017/030}). M. Pereira-Fari\~na is funded by the ``Conseller\'{i}a de Cultura, Educaci\'{o}n e Ordenaci\'{o}n Universitaria'' (under the Postdoctoral Fellowship accreditation \textbf{ED481B 2016/048-0}).

\section*{References}
\bibliographystyle{elsarticle-harv}
\bibliography{bibliography}

\end{document}